\documentclass[sigconf, 9pt]{acmart}

\usepackage{caption}
\usepackage{graphicx}
\usepackage{tabularx} 

\usepackage{makecell} 
\usepackage{xspace}
\usepackage{booktabs}
\usepackage{multirow}
\usepackage{cleveref}
\usepackage{lipsum}
\usepackage[normalem]{ulem}
\usepackage{enumitem}
\usepackage{geometry}
\usepackage{amsmath}
\usepackage{float}
\usepackage{adjustbox}
\setlist[itemize]{noitemsep, topsep=0pt}
\usepackage[utf8]{inputenc}
\usepackage[T1]{fontenc}
\usepackage{adjustbox}
\usepackage{subfigure}
\usepackage[normalem]{ulem}

\usepackage{soul}
\usepackage{hyperref}
\hypersetup{
    colorlinks=true,
    linkcolor=blue,
    filecolor=magenta,
    urlcolor=cyan,
}

\usepackage{siunitx}
\usepackage{tikz}

\makeatletter
\DeclareRobustCommand\onedot{\futurelet\@let@token\@onedot}
\def\@onedot{\ifx\@let@token.\else.\null\fi\xspace}

\makeatother

\AtBeginDocument{%
  \providecommand\BibTeX{{%
    \normalfont B\kern-0.5em{\scshape i\kern-0.25em b}\kern-0.8em\TeX}}}

\newcommand{\system}{\textit{JARVIS}\xspace}

\begin{document}
\title[]{LLM-based Question-Answer Framework for Sensor-driven HVAC System Interaction}


\author{Sungmin Lee$^1$, Minju Kang$^1$, Joonhee Lee$^1$, Seungyong Lee$^2$, Dongju Kim$^2$, Jingi Hong$^2$, \\ Jun Shin$^2$, Pei Zhang$^3$, and JeongGil Ko$^{1,4}$}
\affiliation{%
  \institution{$^1$ School of Integrated Technology, Yonsei University, Seoul, Korea \\ $^2$ LG Electronics, Seoul, Korea \\ $^3$ Department of Electrical and Computer Engineering, University of Michigan, Ann Arbor, MI \\ $^4$ Graduate School of Artificial Intelligence, POSTECH, Pohang, Korea}
  \country{}
}
\begin{abstract}
Question-answering (QA) interfaces powered by large language models (LLMs) present a promising direction for improving interactivity with HVAC system insights, particularly for non-expert users. However, enabling accurate, real-time, and context-aware interactions with HVAC systems introduces unique challenges, including the integration of frequently updated sensor data, domain-specific knowledge grounding, and coherent multi-stage reasoning. In this paper, we present \textbf{\textit{JARVIS}}, a two-stage LLM-based QA framework tailored for sensor data-driven HVAC system interaction. \system employs an \textbf{Expert-LLM} to translate high-level user queries into structured execution instructions, and an \textbf{Agent} that performs SQL-based data retrieval, statistical processing, and final response generation. To address HVAC-specific challenges, \system integrates (1) an adaptive context injection strategy for efficient HVAC and deployment-specific information integration, (2) a parameterized SQL builder and executor to improve data access reliability, and (3) a bottom-up planning scheme to ensure consistency across multi-stage response generation. We evaluate \system using real-world data collected from a commercial HVAC system and a ground truth QA dataset curated by HVAC experts to demonstrate its effectiveness in delivering accurate and interpretable responses across diverse queries. Results show that \system{} consistently outperforms baseline and ablation variants in both automated and user-centered assessments, achieving high response quality and accuracy. 

\end{abstract}

\settopmatter{printacmref=false} 
\renewcommand\footnotetextcopyrightpermission[1]{} 
\setcopyright{none}
\pagestyle{plain}
\maketitle
\section{Introduction}

The sensor-driven segment of the smart building HVAC (Heating, Ventilation, and Air Conditioning) industry, involving applications such as fault diagnosis and energy-efficient control, is projected to represent the most significant area of growth, with an estimated market value of \$8.31 billion by 2029~\cite{thechillbrothers2025hvac}. These systems and their applications typically demand advanced domain expertise and manual effort to query their databases and interpret sensor information. In order to support system interaction with non-expert everyday users, enabling such capabilities through a question-answering (QA) interface offers substantial benefits, providing an interactive, user-friendly, personalized, and demand-responsive experience.

Common approaches for designing such QA applications include employing trained human experts or implementing rule-based bots~\cite{singh2022survey,riloff2000rule}. While human experts can provide accurate and context-aware responses, they incur high labor costs and lack scalability, making them impractical for real-time or continuous support in large-scale deployments. Rule-based automated systems, in contrast, rely on manually crafted templates or decision trees to generate predefined responses. While these systems offer low latency and predictable behavior, they struggle with the variability and ambiguity of natural language queries, particularly when questions span multiple concepts, contexts, or time frames~\cite{hong2024next, jiang2025prompt, lewis2020retrieval, lu2024evaluation,liu2023surveyagent,wang2023rise}. These limitations motivate the potential use of large language models (LLMs), which combine the scalability of automation with the flexibility of natural language understanding. LLMs can generalize across diverse queries and adapt to novel user demands without requiring constant rule updates or expert intervention, making them well-suited for dynamic, sensor-driven domains.

When generating responses, an LLM-based HVAC QA framework must ground its outputs in real-time sensor data produced by HVAC systems. This requirement renders conventional approaches based on statically integrated databases fundamentally unsuitable, as they cannot accommodate the dynamic and frequently updated nature of operational sensor streams. Because LLMs are trained on static corpora and cannot be retrained in real time, the framework must retrieve and inject up-to-date data through alternative methods such as prompt-based augmentation~\cite{lewis2020retrieval,zhao2024retrieval,li2025enhancing,sahoo2025systematic,li2024systematic}.

In practice, HVAC sensor data are stored in structured databases and must be retrieved using database-oriented approaches (e.g., SQL)~\cite{biswal2024text2sql,li2023virtual,rasheed2023leveraging}. Once retrieved, these data commonly require domain-specific preprocessing, such as statistical summarization, temporal filtering, or condition-based selection, to derive semantically meaningful representations suitable for generating a proper response to the user. Such preprocessing is critical for ensuring contextual accuracy of the response, reducing prompt length, and mitigating the LLM's limited capabilities in numerical and logical reasoning~\cite{chen2023unleashing,an2024make,liu2025survey}. This motivates the need for a tightly integrated pipeline that combines structured data retrieval, HVAC-specific data analytics, and natural language-based user interaction to produce coherent and actionable responses.

To support real-time QA interactions for HVAC systems while addressing the aforementioned requirements, we introduce \system, a QA framework tailored for data-driven HVAC applications. \system employs two distinct LLMs in a modular architecture inspired by the master–worker paradigm. Specifically, \system consists of two sequential components: an \textbf{Expert-LLM}, which interprets user requests and generates a set of precise execution instructions grounded in HVAC-specific domain knowledge; and an \textbf{Agent}, which carries out a multi-stage answering process that includes SQL-based data retrieval, Python-based data processing, and LLM-based response generation, based on the Expert-LLM's instructions.

While being a promising approach, designing \system introduces several key challenges. First, translating vague, high-level user requests into precise, low-level execution instructions suitable for the HVAC domain requires the Expert-LLM to incorporate both deep HVAC domain knowledge and deployment-specific context. Effectively integrating such specialized knowledge into general-purpose LLMs, while balancing data requirements and computational cost—remains an open challenge~\cite{lu2024evaluation,zhang2025domain,xu2025reinforcement,fan2024eplus,yang2024heating}. To address this, we classify contextual information based on its versatility and representation length, and adopt an \textit{adaptive context injection} strategy that selects and integrates relevant context per category.

Second, fine-tuning LLMs for SQL query generation is known to be challenging, resource-intensive, and often brittle~\cite{hong2024next,pourreza2023dinsql,zhu2024large}. To overcome this, we introduce a \textit{parameterized SQL builder} that constructs SQL statements from a set of required parameters. This component abstracts away redundant boilerplate syntax, enabling the Expert-LLM to focus on accurately identifying query conditions rather than managing complex SQL syntax.

Note that the multi-stage answering process requires the Expert-LLM to generate structured (and sequential) instructions that carefully guides each subsequent step toward producing a question-relevant final response, while avoiding unnecessary or misaligned intermediate goals. To maintain coherence across consecutive instructions, we adopt a \textit{bottom-up planning} approach, in which the Expert-LLM first defines an expected answer template and then plans preceding the querying and processing steps accordingly. 

We evaluate \system{} using sensor data collected from a real-world, commercial-scale HVAC deployment and a ground truth QA dataset curated by HVAC experts. Our results show that \system{} consistently outperforms ablation variants and baseline configurations across multiple response quality metrics, including cohesiveness, helpfulness, and truthfulness. Through both automated evaluations using an LLM-as-a-Judge and a real-world user study, we demonstrate that each component in \system{} contributes meaningfully to the overall quality of system responses. 

Specifically, we make the following contributions in this work:

\begin{itemize}[leftmargin=*]
    \item We categorize HVAC-specific and deployment-specific contextual information required for QA applications, and propose an adaptive context injection strategy for data- and cost-efficient integration into general-purpose LLMs.

    \item We achieve LLM-driven sensor data retrieval through a parameterized SQL builder and executor that leverages template-based query construction, to reduce fragility and improve reliability.

    \item We introduce \system, a two-stage LLM-based QA framework for data-driven HVAC applications, leveraging bottom-up planning and variable alignment to ensure coherent multi-step execution.

    \item We evaluate \system using sensor data collected from real-world, building-scale HVAC deployments and an expert-curated QA dataset to show its effectiveness as an automated QA system.
\end{itemize}


\section{Background and Related Work}
\label{sec:background}

\subsection{LLM in HVAC systems}
\label{subsec:hvacexpertLLM}

Recent studies have begun to explore the use of large language models (LLMs) in HVAC applications. Lu et al.~\cite{lu2024evaluation} evaluated commercial LLMs such as GPT-4 and LLaMA for their HVAC-related knowledge and reasoning capabilities using the ASHRAE Certified HVAC Designer examination~\cite{ashrae_chd}, a professional certification used to assess HVAC expertise across a wide range of applications. Their results revealed that general-purpose LLMs lack sufficient domain knowledge, suggesting the need to inject HVAC-specific knowledge and application context into LLMs during inference.

Prior works have primarily adopted two approaches for incorporating such domain knowledge. Zhang et al.~\cite{zhang2025domain} developed an HVAC fault-diagnosis assistant by fine-tuning a general-purpose LLM using labeled sensor data from building systems. As emphasized in their work and in other LLM literature~\cite{lewis2020retrieval,zhao2024retrieval,su2024dragin,acm2024seven,he2024gretriever}, fine-tuning integrates rich and complex knowledge directly into model weights. This approach supports the reliable injection of large context volumes, but lacks adaptability to post-deployment updates and requires a costly, domain-specific training dataset. 

Alternatively, some studies have employed prompt engineering, which involves prepending relevant contextual information to the original query at inference time. Jiang et al.~\cite{jiang2025prompt} utilized prompt engineering to construct an LLM-based assistant for automated building energy modeling. This approach enables rapid prototyping and the incorporation of up-to-date information, but typically suffers from higher inference latency and reduced accuracy when processing complex knowledge or long-context inputs~\cite{an2024make,chen2023unleashing,levy2024same,sahoo2025systematic}.

Overall, these prior work suggest that building an HVAC-expert LLM requires a balanced strategy that leverages the strengths of both fine-tuning and prompt engineering to meet the demands of domain specificity, flexibility, and inference efficiency.

\subsection{Incorporating Up-to-Date Time-Series Sensor Data from Tabular Databases}

User-interactive HVAC applications rely on leveraging real-time time-series sensor data collected from building sensors and (often) stored in databases. Since such data are not accessible during the fine-tuning stage for real-world deployments, LLMs for these applications require an external framework that retrieves relevant sensor data at inference time and augments them via prompting (e.g., retrieval-augmented generation (RAG))~\cite{lewis2020retrieval,ning2025tsrag,zhao2024retrieval,gao2023retrieval,fan2024survey}.

As noted in Section~\ref{subsec:hvacexpertLLM}, 
LLM performance tends to degrade and inference cost increases with longer inputs. Therefore, it is crucial to only exploit data that is both highly relevant and concise in the augmentation process, avoiding unnecessary or irrelevant information~\cite{levy2024same}. While some previous work such as TaskSense attempt to access external data via external API calls~\cite{liu2025tasksense}, a more general approach for augmenting LLMs with time-series sensor data stored in relational databases would involve translating user input into SQL queries (text-to-SQL~\cite{yu2018spider,pourreza2023dinsql,zhu2024large,hong2024next}) to identify and leverage the most relevant sensor information. Biswal et al.~\cite{biswal2024text2sql} introduced a two-stage LLM architecture, employing a front-end LLM to perform text-to-SQL translation and a back-end LLM to generate natural language responses based on the retrieved data.

Our work acknowledges the observations made by Biswal et al., and targets to   address several key limitations and challenges. First, the LLMs that Biswal et al. use for both stages are general-purpose LLMs. Thus, their system lacks HVAC-specific domain knowledge and therefore is likely to underperform in HVAC-specific scenarios. We address this by injecting HVAC-specific and deployment-context knowledge to the front-end model, enabling it to relay essential domain information to the general-purpose back-end model for more accurate answer generation.

Second, our preliminary experiments reveal that directly generating complete SQL statements (i.e., boilerplate syntax, nested queries, and multiple joins) negatively impacts the system robustness due to inconsistencies and errors in the generated queries themselves. To mitigate this, we propose the parameterized SQL builder where the LLM produces an abstract query plan or key-value directives, and a rule-based module then completes the final SQL statement, ensuring syntactic and semantic correctness.

Third, general-purpose LLMs typically lack the mathematical and logical reasoning capabilities required to analyze raw HVAC sensor data. To overcome this, we introduce a Python-based data processing pipeline between the retrieval and answer generation stages similar to the approach taken by Xiao et al.~\cite{xiao2024exploring}. This module performs necessary computations, such as statistical summarization and formatting, to enhance the LLM reasoning process and reduce prompt complexity for response generation.

\subsection{Preliminary study: Failure analysis of text-to-SQL in time-series HVAC data}
\label{subsec:prelim}

\begin{figure}[!t]
    \centering
     \subfigure[Query success rate with respect to inputs taxonomy]{
    \includegraphics[width=0.47\linewidth]{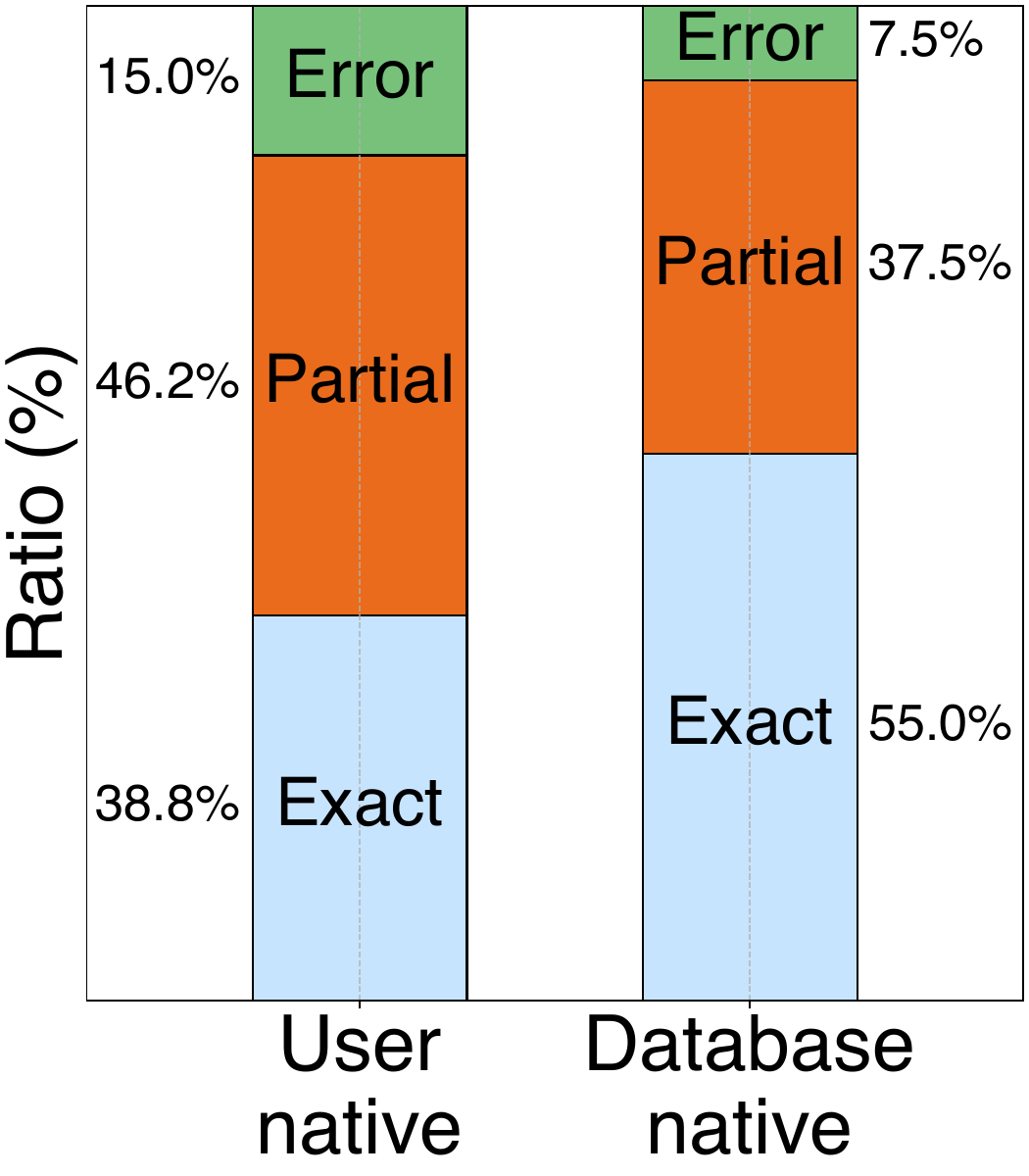}
    \Description{}
    }
    \subfigure[Reasons of failure (partial + error) for database-native inputs]{
    \includegraphics[width=0.48\linewidth]{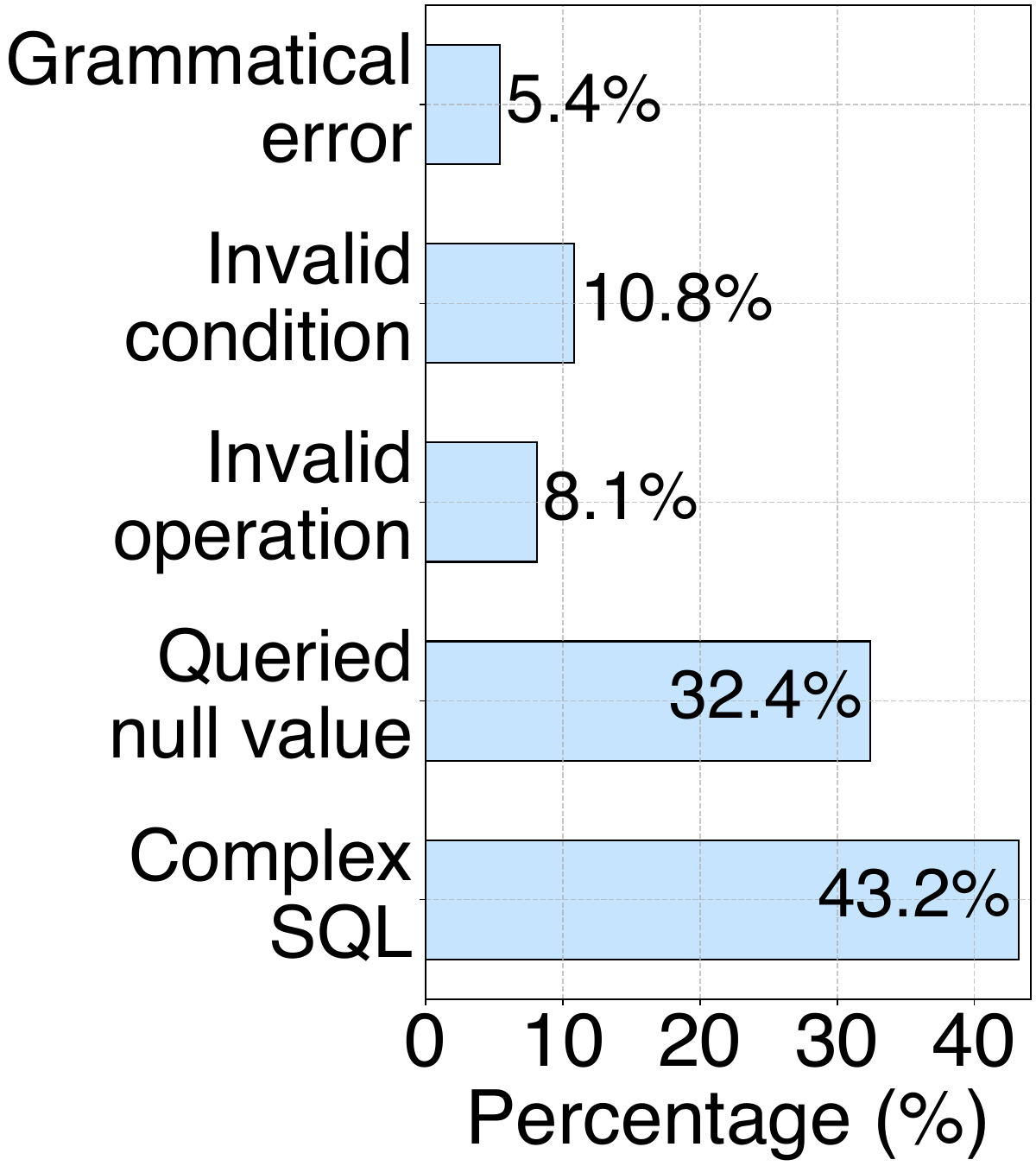}
    \Description{}
    }
    \vspace{-3ex}
    \caption{Failure analysis of generic text-to-sql operations in an HVAC scenario.}
    \vspace{-2ex}
    \label{fig:prelim-sql}
    \Description{}
\end{figure}

In this section, we analyze failure cases of text-to-SQL operations to better understand the challenges discussed earlier and motivate the design choices in our proposed approach. The analysis was conducted using a simple toy configuration as follows. We used one year of HVAC time-series sensor data collected from three rooms from a real-world large-scale building deployment, covering two modalities: room temperature and user-desired temperature. We note that this data is collected from a commercial deployment and we have gained the proper permissions to analyze and publish analytical results using the data. Based on this time-series data, HVAC experts curated 80 question-SQL pairs spanning a range of user requests, from simple queries (e.g., retrieving the current temperature of a room) to more complex statistical ones (e.g., identifying the date last month when the daily average temperature was lowest).  

To assess the role of language abstraction, each question (i.e., LLM query) was prepared in two taxonomic forms. First, user-native queries used natural phrasing familiar to non-expert users (e.g., \textit{``How hot is our room?''}), while database-native queries directly aligned with the schema without ambiguous or missing conditions (e.g., \textit{``What is the current roomtemp of room101?''}, where \texttt{roomtemp} and \texttt{room101} correspond to column names and identifiers in the database). A LLaMA3-8B model, fine-tuned for general-purpose text-to-SQL, was used as the baseline model for this purpose. 

Using this setup, we perform two different experiments. In the first experiment, we evaluated the model's ability to handle user-native taxonomies.  
For each user request, both the user-native and corresponding database-native version were fed into the model, and the generated SQL was compared with the expert-crafted ground truth by checking if the query results matched exactly (exact match). Ideally, the LLM should output SQL statements semantically similar to the ground truth. In case a mismatch occurs, we classify it as a partial match (including cases with no overlap). If execution failed due to syntax errors or invalid structure, we labeled it as an error.  
When evaluating user-native inputs, we also provided the model with the mapping between user-native and database-native taxonomies as part of the system prompt.  

As shown in Figure~\ref{fig:prelim-sql}~(a), the model performed significantly better on database-native queries, where queries were ``intuitive'' to understand.  
The exact match rate for user-native inputs remained below 40\%, and the gap in performance suggests that even with explicit taxonomy mapping, the general-purpose text-to-SQL model struggles to bridge the abstraction gap.  
This suggests the need for improved approaches to interpreting natural, user-native inquiries.

In the second experiment, we investigated the causes of failure by asking four HVAC experts to debug and identify the erroneous outputs (Figure~\ref{fig:prelim-sql}~(b)).  
The most common source of failure was the generation of fragile and overly complex SQL statements, particularly those involving subqueries and multi-table joins.  
These queries are prone to syntactic and logical errors, especially since many text-to-SQL models are pre-optimized to generate a single comprehensive SQL per request.  
For statistical operations such as \texttt{argmax}, which are not natively supported in SQL, the model attempts to simulate the operation using complex constructs, further increasing the likelihood of failure.

Another significant failure cause we noticed was the improper handling of missing sensor values. As is common in many real-world HVAC deployments~\cite{Cheng2019HVAC,li2020handling}, our dataset contained a fairly large number of null entries representing sensor failures. In these preliminary experiments we instructed the model via prompt to exclude these values during statistical computation using appropriate SQL filtering (e.g., \texttt{IS NOT NULL}). Nevertheless, the model frequently failed to apply such conditions reliably.

Based on these observations from our preliminary study, we focus on two major design changes.  For handling missing values, we use post-generation boilerplate logic to automatically append filtering conditions that exclude null elements, improving robustness without burdening the model. To address the issue of complex SQLs, rather than relying on the model to generate a monolithic query, we instead train a custom LLM to generate a sequence of short, modular SQLs that can be combined to generate a final result. This approach not only reduces the chance of syntax errors but also increases reusability and consistency, as common query patterns emerge across simpler blocks. Together, these solutions motivate a parameterized SQL generation framework, where the model is responsible for specifying high-level query intentions and leaves repetitive or syntax-heavy component (e.g., joins, filters) generation to deterministic logic.

Finally, we also address the common failure regarding the incorrect use of SQL's built-in statistical functions. To support operations not available in SQL or prone to misinterpretation, we use Python-based post-processing within the system pipeline to handle such computations. This integrated pipeline architecture ensures accurate statistical analysis and expands the range of supported user queries and operations.

\begin{figure*}[t]
    \centering
    \includegraphics[width=.9\linewidth]{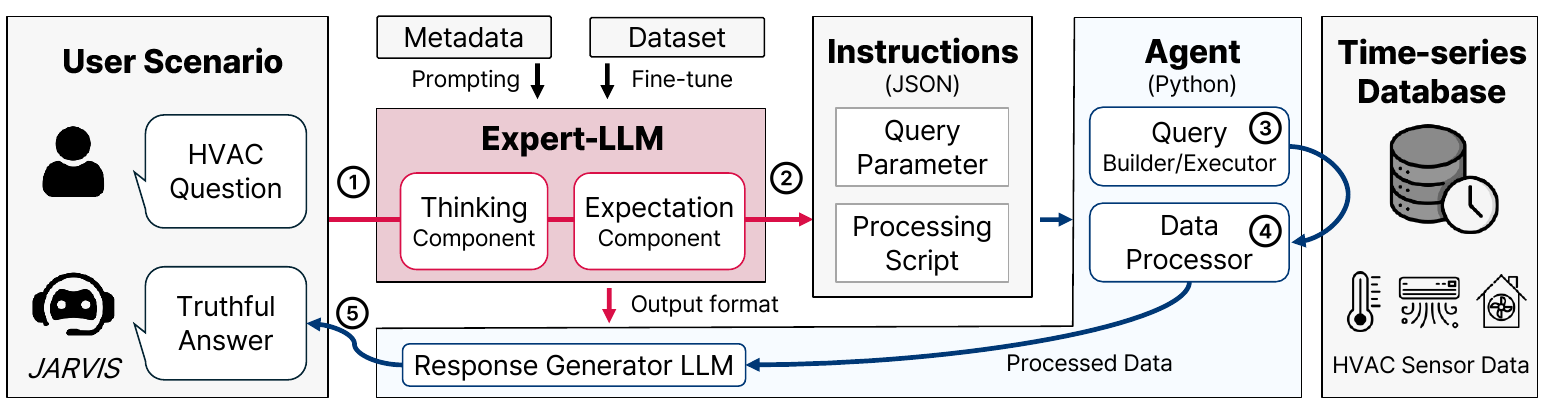}
    \vspace{-2ex}
    \caption{Operational flow of \system{}.}
    \vspace{-2ex}
    \label{fig:framework_overview}
\end{figure*}

\begin{table*}[]
\centering
\begin{adjustbox}{width=\linewidth} 
\begin{tabular}{@{}l|cc|cc>{\raggedright\arraybackslash}p{0.18\linewidth}>{\raggedright\arraybackslash}p{0.18\linewidth}>{\raggedright\arraybackslash}p{0.18\linewidth}>{\raggedright\arraybackslash}p{0.18\linewidth}@{}}
 & \multicolumn{2}{c}{\textbf{Semantic context}}& \multicolumn{2}{c}{\textbf{Sensor context}}\\
\toprule
                    & \textbf{HVAC-common knowledge}   & \textbf{Deployment and 
user specific context} & \textbf{Small-scale sensor data} & \textbf{Large-scale sensor data} \\ \midrule \midrule
Occurrence time & Before deployment         & Before request                        & In-situ (real-time)                 & In-situ (real-time)                 \\ \midrule
Context length& Long& Short                          & Short                   & Long                    \\ \midrule \midrule
\textbf{Pretraining}         & X                         & X                              & X                       & X                       \\
\textbf{Finetuning}          & O                         & X                              & X                       & X                       \\
\textbf{Direct Prompting}       & -                         & O                              & O                       & X                       \\
\textbf{Processed Prompting} & -                         & -                              & O                       & O                       \\ \midrule\midrule
Examples             & Presuppositions & Per-user preference                & Short-term data         & Long-term data          \\ \bottomrule
\end{tabular}
\end{adjustbox}
\caption{Ideal/target context injection method with respect to contextual characteristics in a domain-specific QA system (`-': Not necessary, `O': Applicable, `X': Not applicable).}
    \label{tab:injection_method}
    \vspace{-2ex}

\end{table*}

\section{\system{} Design}

\subsection{Design Overview}

As an LLM-based HVAC QA framework, \system{} adopts a master-worker architecture with a five-stage answering process, as depicted in Figure~\ref{fig:framework_overview}. Specifically, \system{} utilizes an Expert-LLM (master) to interpret user input and orchestrate the overall answering pipeline (\textcircled{\small{1}}). Specifically, this Expert-LLM generates a set of \textit{execution instructions} derived from the user request (\textcircled{\small{2}}), which layout the tasks required to generate a user response. Based on these instructions, \system's Agent (worker) performs three core operations: (i) querying real-time sensor data from the database (\textcircled{\small{3}}), (ii) processing the queried data (\textcircled{\small{4}}), and (iii) generating a natural language response (\textcircled{\small{5}}). Note that the final response is generated by a general-purpose LLM, guided by both the retrieved and processed data as well as a pre-defined set of answering guidelines tailored to the application.



\subsection{Adaptive Context Injection}
\label{subsec:adaptivecontext}

As discussed in Section~\ref{sec:background}, incorporating contextual information into LLMs is essential for effective HVAC QA, but prior work typically relies on a single injection method (i.e., prompting or fine-tuning) without proper justification. In contrast, we classify the context into four types based on versatility and representation length, and apply an \textit{adaptive injection} strategy tailored to each category (Table~\ref{tab:injection_method}).



\vspace{1ex}
\noindent \textbf{HVAC-common knowledge}, such as domain-specific reasoning patterns (e.g., time-series data analysis) or linguistic conventions in user queries, is deployment-agnostic and stable across scenarios. This type of context often requires lengthy descriptions and diverse examples, making it best suited for fine-tuning. Once embedded into model weights, this knowledge remains robust after deployment.

\vspace{1ex}
\noindent \textbf{Deployment-specific and user-specific contexts}, including sensor layout maps or individual user preferences, are dynamic and vary by installation. These inputs are typically compact (e.g., configuration files or metadata) and can be efficiently injected via prompting at inference time.

\vspace{1ex}
\noindent \textbf{Small-scale sensor data} from short time-series windows, are concise and adaptable, making them ideal for direct prompting. However, \textbf{large-scale sensor data}, such as high-frequency or multi-day streams, can exceed the input context length. Additionally, LLMs often struggle to perform necessary statistical analysis or anomaly detection directly on raw data. For this reason, we propose to algorithmically preprocess or summarize such data prior to injection.

\vspace{1ex}
Overall, \system{} adopts this \textbf{adaptive context injection} strategy by categorizing input contexts according to the above framework and selecting the optimal injection method to balance answer quality and inference cost. In particular, we fine-tune the expert front-end LLM with HVAC-common knowledge and dynamically provide it with deployment- and user-specific contexts via prompting. For sensor data, we follow the execution instructions generated by the expert-LLM to retrieve and process relevant subsets. Large-scale sensor data are algorithmically summarized before being passed as input, thus mitigating prompt length limitations while preserving essential information.

Throughout the paper, we will refer to the HVAC-common knowledge and deployment- and user-specific contexts as \textit{semantic context} and the both sensor data types as \textit{sensor context}.

\begin{figure*}[t!]
    \centering
    \includegraphics[width=.9\linewidth]{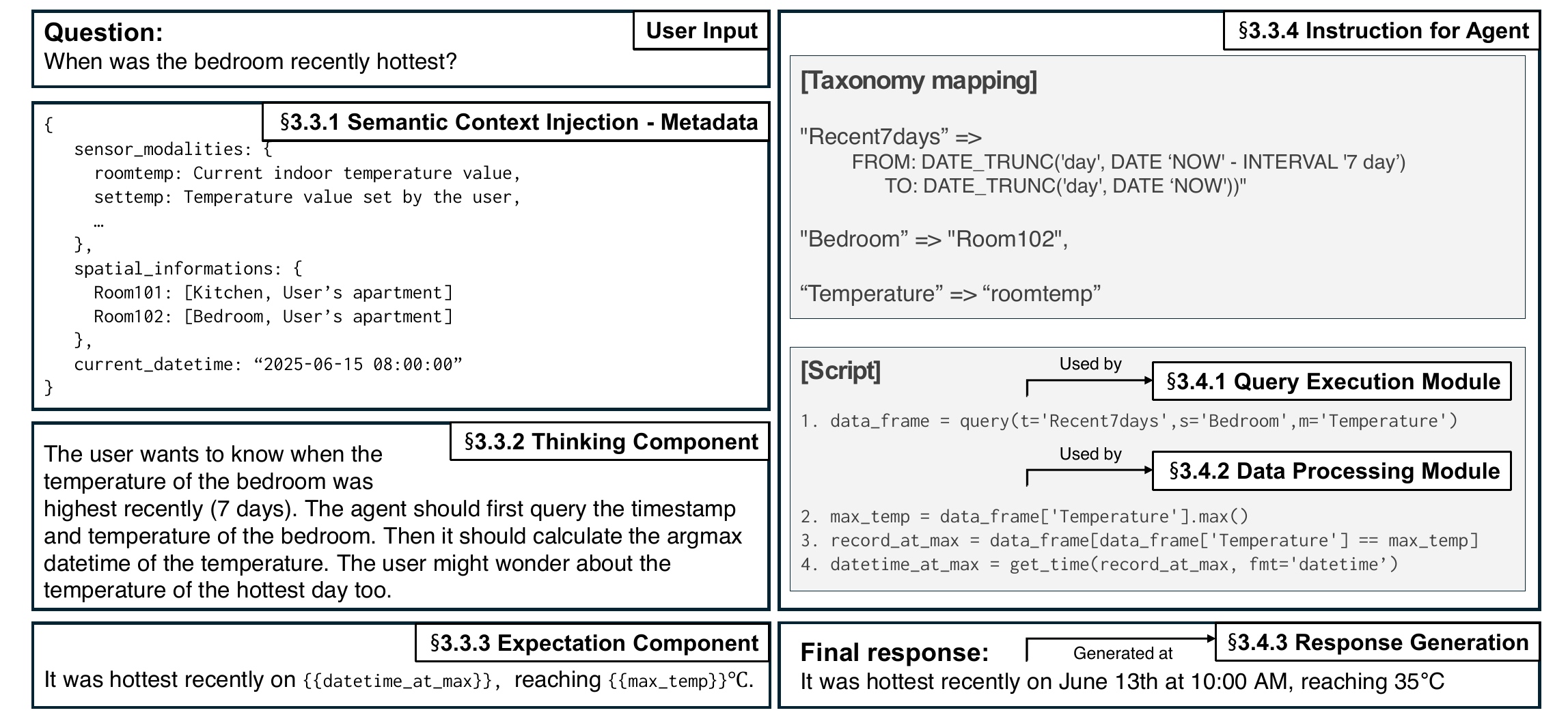}
    \vspace{-2ex}
    \caption{Sample operation of Expert-LLM generating execution instructions based on user input.}
    \vspace{-2ex}
    \label{fig:example}
\end{figure*}

\subsection{Expert-LLM}
\label{subsec:expertLLM}

As demonstrated in our preliminary study and through prior work, using a text-to-SQL model alone to understand user requests and generate executable queries is insufficient for HVAC applications.  
Instead of proposing incremental improvements to existing text-to-SQL approaches, \system{} introduces a specialized HVAC-expert LLM, built on top of a general-purpose LLM, and trained to serve a broader and more semantically rich role.

Unlike traditional text-to-SQL models, which focus solely on converting natural language into executable SQL statements, the HVAC-expert LLM acts as a translator between the user and the system agent. Specifically, its responsibilities are twofold:  
(1) to incorporate \textit{semantic context} and accurately interpret user-native requests; and  
(2) to plan and direct the system's response pipeline by generating structured, executable instructions.

The operation of the \textbf{Expert-LLM} begins with a natural language query provided by the user. Upon receiving the query, the model first interprets its meaning in the context of relevant semantic and contextual information (e.g., HVAC-specific terminology, prior dialogue state). It then produces a set of structured execution instructions, formatted in JSON, which guide the downstream components, including data retrieval, processing, and response generation, toward producing an accurate and context-aware answer.

\subsubsection{Semantic Context Injection}

To enable the HVAC-expert LLM to interpret user-native queries and plan appropriate actions, semantic context must be integrated (Section~\ref{subsec:hvacexpertLLM}). In \system{} we employ both prompting and fine-tuning for this context injection process.

For deployment- and user-specific contexts, we use prompting at inference time. Specifically, in \system{}, these contexts are structured as a key-value dictionary referred to as \textit{metadata}, comprising two core components:  
(1) a mapping between user-native and database-native taxonomies for spatial information and sensing modalities, and  
(2) inference-time environmental factors such as the current date and time.  
We note that this structure is extensible and can be customized based on application-specific needs.

For HVAC-common knowledge, we fine-tune the base LLM using an expert-curated dataset. While the scope of this knowledge may vary across deployments, we identify a set of generalizable knowledge components that are broadly applicable, such as presuppositions underlying HVAC-related queries, preferred answering strategies, typical data query patterns, and standard data processing procedures. This process also teaches the Expert-LLM to produce well-structured, syntactically valid JSON instructions.

\subsubsection{Thinking component: Intermediate thought generation}

Prompting typically imparts static rules and constraints to an LLM, whereas fine-tuning enables the model to emulate expert behavior, capturing both explicit patterns and implicit reasoning strategies. Prior work has shown that generating intermediate reasoning steps via the chain-of-thought (CoT) approach, can significantly enhance the reasoning capabilities of autoregressive LLMs~\cite{wei2022chain,zhang2024spiqa,chu2023survey,gupta2024comprehensive,asai2023retrival}. This not only improves inference accuracy but also increases transparency by exposing the model’s decision-making process.

To leverage these benefits, \system{} incorporates a dedicated \textbf{Thinking Component} that guides the Expert-LLM to explicitly emulate human-like reasoning through intermediate ``thoughts'' prior to instruction generation. A human expert answering an HVAC-related query typically interprets the question within its context, identifies the relevant scope of data and tools, develops a high-level plan, and finally converts this plan into actionable instructions. In \system{}, the thinking component addresses the first three stages, while the final translation into structured instructions is handled by the instruction generation module (see Sec.~\ref{sec:instuction-gen}).

Within the thinking component, the Expert-LLM focuses on three core functions. First, it resolves ambiguous or user-native terminology by aligning it with the database-native taxonomy defined in the metadata. Second, it infers and fills in missing or underspecified conditions based on presuppositions learned during fine-tuning, resulting in a more explicit and actionable reformulation of the user query. Third, it outlines a plan for answering the query, including necessary data retrieval and processing steps, to ensure all relevant information is collected before response generation begins.

To support these capabilities, we manually curate training examples using the CoT methodology, capturing each of these reasoning processes. We present an example of such training in Figure~\ref{fig:example}.

\subsubsection{Expectation Prediction and Bottom-up Planning}

Upon generation of the Thinking component, the Expert-LLM produces the \textbf{User Expectation Component}, serving three primary purposes.

First, the User Expectation component anticipates additional information that the user may implicitly expect. For example, given the request \textit{``When was our room coldest today?''}, the Expert-LLM infers that the user likely also wants to know the actual temperature at that time. By predicting such latent expectations based on HVAC-common knowledge, the system enriches the user experience by proactively including useful supplemental details.

Second, it guides the response generation LLM with a context-aware answer format. Even when the same data is available, the preferred tone, level of detail, and structure of the response can vary depending on HVAC norms or individual user preferences. The user expectation component provides an explicit example format that the response generator can follow, serving as a direct and lightweight injection mechanism to convey contextual instructions.

Third, it defines the data that must be queried and processed by the agent before initiating answer generation. For instance, in response to the query \textit{``Has the temperature in Room 101 been unstable recently?''}, the Expert-LLM may set the expectation to include a time window and a summary of daily fluctuations; thus, enabling the system to prepare all necessary data in advance.

By explicitly defining and training samples of these expectations, \system{} performs bottom-up planning as it establishes the final output goal and works backward to ensure that all necessary upstream querying and processing steps are included. We provide an example of such a user expectation component training sample in Figure~\ref{fig:example}.


\subsubsection{Generating Instructions for the Agent}
\label{sec:instuction-gen}

After completing the thinking and expectation components, the Expert-LLM proceeds to generate executable instructions for the agent. As the example in Figure~\ref{fig:example} shows, these instructions are structured in JSON format, allowing the agent to parse and execute them deterministically. Note that the reliability of \system{} partially depends on the Expert-LLM's ability to generate syntactically correct and semantically meaningful JSON. To ensure this, we combine fine-tuning with lightweight prompt-based guidance, which together enable the model to produce consistently well-formed outputs. 

The specific structure and contents of these instructions will be detailed in the following sections, alongside an introduction to the Agent and its role in the overall execution pipeline.

\subsection{Agent}

Given the execution instructions generated by the Expert-LLM, the Agent operates on top of a Python runtime to perform the specified tasks. In \system{}, the Agent is responsible for three primary operations: data querying, data processing, and response generation.  

For data querying and processing, the instructions follow a simple imperative structure in the form of \texttt{variable = expression}, where the right-hand side expression is evaluated, and its result is stored in the specified variable for use in subsequent operations. For response generation, a general-purpose LLM is invoked to ensure the final user-facing response is both contextually grounded and semantically fluent.

\subsubsection{Query Execution Module}

As highlighted in the preliminary study (Section~\ref{subsec:prelim}), conventional text-to-SQL approaches often fail in HVAC applications due to three key issues:  
(1) inability to handle user-native taxonomies,  
(2) brittleness when generating complex, monolithic SQL statements, and  
(3) failure to include necessary boilerplate syntax, such as null-value filtering.

To address these limitations, we introduce a \textbf{Parameterized Query Builder/Executor} approach. In \system{}, the Expert-LLM is only responsible for expressing high-level query intentions using user-native terminology (c.f., Fig.~\ref{fig:example}), while a rule-based SQL generation module (running within the Agent) converts these into complete, executable SQL statements, including complex constructs and boilerplate logic.

By decoupling the high-level intent from the low-level SQL generation, we overcome limitations of traditional text-to-SQL models, such as the constraint of producing a single query per inference and the associated latency and fragility of generating deeply nested SQL. Instead, the expert LLM is free to issue multiple modular query instructions, improving interpretability and robustness.

Each instruction to the parameterized query builder consists of two components: \textit{mapping} and \textit{calling}. The mapping defines a key-value dictionary that aligns user-native taxonomy (refined during the thinking stage) with database-native taxonomy (provided via the metadata). The calling component triggers a function within the Agent’s runtime that performs SQL generation and querying.

This function uses the user-level conditions specified in the instruction and applies the provided mapping to convert them into database-native fields. It then constructs the raw SQL query, including any necessary subqueries, joins, or filtering clauses, and executes it against the database. To support extensibility, complex operations can be invoked using high-level keywords, which are internally expanded into SQL patterns.  

The queried result is returned as a Pandas DataFrame~\cite{mckinney2010data,reback2020pandas}. While the raw output is in terms of database-native taxonomy, the function automatically converts the column labels back into user-native terminology before returning it, ensuring downstream compatibility with the rest of the pipeline. Note that, this calling process can be executed multiple times, allowing for modular, composable query operations.

In summary, our parameterized SQL builder and executor design enhances the robustness and modularity of the data access pipeline. By abstracting away the complexities of SQL generation (i.e., schema knowledge, boilerplate syntax, and error handling) we enable the expert LLM to focus purely on producing high-level, natural-language-aligned instructions. Furthermore, this functionalized approach offers greater system control: for example, a data-security middleware could intercept and vet queries before execution, preventing unauthorized access to sensitive information.

\subsubsection{Processing Queried Data}
Once the data is queried, \system{}'s \textbf{Data Processor} processes the retrieved data to extract meaningful insights based on the Expert-LLM's instructions (c.f., Fig.~\ref{fig:example}). This design mitigates the limitations of LLMs in handling mathematical reasoning and also reduces the cost of response generation. In \system{}, we use Python's Pandas library~\cite{reback2020pandas} for data processing rather than relying on native SQL operations, which often lack the necessary functionality. Notably, more complex or HVAC-specific processing, such as diagnostic modules or classifiers, can also be integrated at this stage, provided they are implemented in Python and conform to the expected input-output interface.

\subsubsection{Response Generation Using a General-Purpose LLM}

Leveraging the user's original request, the summarized and processed data, and the guidance provided by the User Expectation component, \system{} leverages a general-purpose LLM to generate the final response presented to the user.
Here, the Expectation component plays a central role in shaping the tone, structure, and completeness of the output, serving as a soft template or reference for the LLM.  

If any required data is missing, either due to execution failure or unavailable sensor readings, the LLM is prompted to generate a graceful fallback response.  
This may include a natural-language explanation of the partial result or an error message clarifying the reason for failure, guided by the expectation context.
This flexibility ensures that the system remains robust and communicative, even when operating under uncertainty or incomplete data availability.

\section{Implementation Details}

This section details the implementation of \system{}. For the Expert-LLM, we use the {\tt LLaMA3.1-8B-Instruct} model~\cite{grattafiori2024llama3herdmodels}, fine-tuned using low-rank adaptation (LoRA)~\cite{hu2021loralowrankadaptationlarge} on a locally curated HVAC dataset consisting of 53 expert-authored samples in the target deployment language (among 80 samples). As the response generation LLM, we employ the {\tt LG EXAONE-3.5-7.8B} LLM~\cite{research2024exaone35serieslarge} without fine-tuning, selected for its strong performance in producing high-quality natural language responses in the target deployment language. Each model is guided by a short prompt that specifies the desired output format and structure. For time-series HVAC sensor data storage, we deployed an instance of the TimescaleDB~\cite{timescaledb}, which is well-suited for managing large volumes of real-time sensor data. The data processing module in \system{} is implemented using the Pandas library atop a Python runtime.

For the evaluations that follow, we deploy \system{} on a server equipped with an Intel Xeon Gold 6448Y 128-core CPU with 128\,GB of RAM, and a single NVIDIA H100 SVM GPU (80\,GB VRAM). The time-series database is hosted on a separate server (on-site) running an AMD EPYC 7443P 24-core CPU, 128\,GB of RAM, and a 2\,TB SSD.

\section{Evaluation}

We evaluate \system{} using data from a real-world, commercial-scale HVAC deployment at a large building site. Based on this data, we constructed a QA dataset curated by HVAC experts, which serves as the foundation for our evaluation (details in Section~\ref{sec:qadataset}).

Our evaluation combines two complementary methods. First, we use the \textit{LLM-as-a-Judge} framework, where a separate LLM assesses whether \system{}'s responses meet the expectations defined in the dataset, enabling scalable, automated evaluation. Second, we conduct a user study with 22 participants, who rate the cohesiveness, helpfulness, and truthfulness of responses to provide human-centered insights into system quality.

\subsection{HVAC QA Dataset Collection}
\label{sec:qadataset}

\subsubsection{HVAC Data Collection Environment}

We collect and utilize one year of HVAC time-series sensor data, sampled at one-minute intervals, from a real-world, large-scale building deployment consisting of 156 units\footnote{This data is collected from a commercial deployment. We have obtained the necessary permissions to analyze the data and publish results derived from it.}. All sensor readings are stored in a tabular database to support structured querying and efficient retrieval.

To simulate realistic user interactions and evaluate the system’s flexibility, we define four user personas: two non-expert residents and two HVAC-expert building managers. Non-expert residents are designed to query only their own HVAC units, using highly abstract and user-native language. In contrast, expert building managers issue queries that focus on building-wide statistics, expressed in explicit and domain-specific terms. Each persona is associated with a distinct user-specific context (i.e., metadata), and queries span a range of time frames to reflect diverse informational needs.


\begin{figure}[!t]
    \centering
    \includegraphics[width=.999\linewidth]{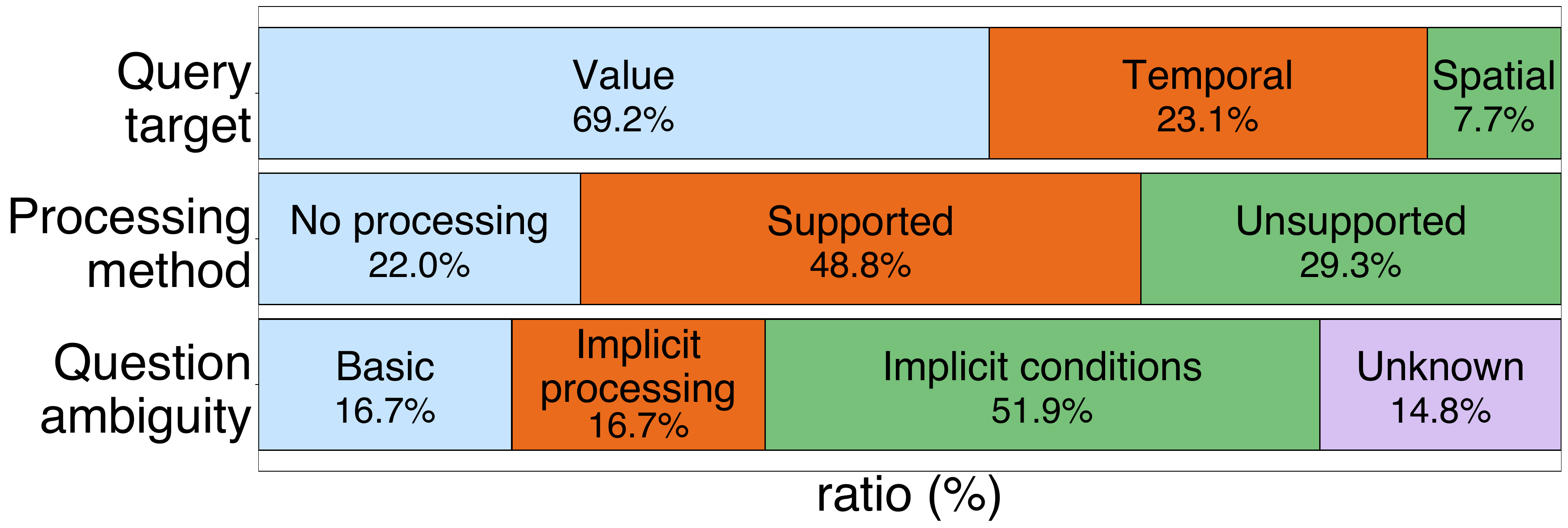}
    \vspace{-4ex}
    \caption{Query type distribution for QA dataset.}
    \vspace{-2ex}
    \label{fig:data-distribution}
\end{figure}

\subsubsection{QA Dataset Curation}

To support both fine-tuning and evaluation of \system{}, human experts curated 80 question–answer (QA) pairs spanning four user personas and a broad range of HVAC-related applications. For each QA pair, HVAC experts authored reference answers based on domain knowledge and manual querying of the sensor data. These answers were then converted into system-specific formats for \system{} and baseline models. In the case of \system{}, each QA pair was decomposed into its full set of expert LLM components: \textit{thinking}, \textit{expectation}, and \textit{instruction}.

We allocated one resident persona and one manager persona to the fine-tuning set, with the remaining two assigned to the test set. The fine-tuning set includes 34 and 19 QA pairs from the resident and manager persona, respectively, while the test set contains 17 and 10, yielding an approximate 2:1 split. Note that, the metadata used for fine-tuning and test personas is disjoint, enabling us to evaluate \system{}'s zero-shot generalization to new deployment and user contexts. The dataset was constructed in the local language.

As Figure~\ref{fig:data-distribution} shows, user queries are well-distributed along three dimensions: \textit{question ambiguity}, \textit{processing method}, and \textit{query target}. A single query may belong to multiple categories within each dimension, as determined by the experts who created the dataset.

\vspace{1ex}
\noindent \textbf{Question ambiguity} captures how implicitly a query conveys its intended operation. Queries are labeled as involving \textit{implicit processing} or \textit{implicit conditions} when they omit explicit phrasing of data operations or constraints, especially when expressed using non-HVAC or colloquial terms. Here, \textit{Unknown} queries refer to requests involving nonexistent data (e.g., invalid time ranges, missing modalities, or unmonitored spaces). In such cases, \system{} is expected to return a truthful failure message rather than hallucinate an answer. This distribution demonstrates that the dataset includes diverse ambiguous and underspecified queries to test the Expert-LLM's semantic grounding capabilities.

\vspace{1ex}
\noindent \textbf{Processing method} indicates the level of data processing needed to answer the query. \textit{No processing} denotes direct retrieval of stored values; \textit{supported} refers to operations that can be handled using native SQL (e.g., filtering or aggregation); and \textit{unsupported} denotes operations requiring additional computation (e.g., \texttt{argmin}, \texttt{argmax}) that must be executed in the agent’s Python runtime. This classification highlights the need for complex processing beyond standard database capabilities in typical HVAC applications.

\vspace{1ex}
\noindent \textbf{Query target} specifies the type of information requested. \textit{Value} targets raw sensor readings; \textit{temporal} includes time-based queries (e.g., ``when was...''), and \textit{spatial} covers location-based queries, typically relevant to building managers. This diversity allows comprehensive evaluation of the system’s ability to handle a range of reasoning and retrieval tasks.



\subsection{Evaluation Metrics}

Evaluating the performance of LLM-based systems on real-world data is inherently challenging due to the limited availability of large-scale labeled datasets and the subjective nature of natural language interpretation. To address this, we define a set of both quantitative and qualitative metrics to assess the performance of \system{}.  These metrics are designed to capture different aspects of response quality, and we describe them in detail below.

\vspace{1ex}
\noindent \textbf{Data querying accuracy metrics.} The accuracy of data query operations are evaluated using three widely-used cell-level (i.e., (row, col)) metrics. These metrics evaluate the accuracy of the SQL statements generated by \system{}, which involves successful operations at the Expert-LLM and the Query Middleware at the Agent.

\begin{itemize}[leftmargin=*]
    \item \textbf{Execution Accuracy~\cite{yu2018spider}}: True if the queried cell exactly matches the ground truth query, else false.
    \item \textbf{Precision}: Proportion of queried cells that are actually correct.
    \item \textbf{Recall}: Proportion of correct cells among all queried cells.
\end{itemize}

We note that in reporting these values, we aggregate each metric by taking the average values per query.

\vspace{1ex}
\noindent \textbf{Response quality metrics.} The quality of the final responses generated by \system{} is evaluated qualitatively.  Following prior work~\cite{ouyang2022training,li2024llms,liu2024behonest,zhang2021vica,gupta2024userstudy}, we adopt three metrics that assess different dimensions of response quality, each rated on a 1–5 Likert scale:

\begin{itemize}[leftmargin=*]
    \item \textbf{Question-Answer Cohesiveness~\cite{ouyang2022training,li2024llms,zhang2021vica,gupta2024userstudy}} measures the semantic alignment between the user’s question and the generated answer (i.e., how closely the content of the answer pertains to the subject and intent of the question).
    
    \item \textbf{Helpfulness~\cite{ouyang2022training,li2024llms,zhang2021vica,gupta2024userstudy}} assesses the extent to which the answer delivers clear and self-contained information that requires minimal user interpretation or reasoning, as well as the presence of relevant auxiliary information that, while not explicitly requested, may enhance user understanding or utility.
    
    \item \textbf{Truthfulness~\cite{ouyang2022training,li2024llms,liu2024behonest,zhang2021vica}} evaluates the factual accuracy of the response compared to a given ground truth, particularly for verifiable elements such as numerical values, factual claims, or explanatory statements.
\end{itemize}

Throughout the evaluation. excluding the user study, we assess response quality using the \textit{LLM-as-a-Judge} methodology~\cite{li2024llms,zheng2023judging,zhang2024spiqa,chen2025generation,liu2023surveyagent}. In this approach, external LLMs serve as surrogate evaluators. Specifically, we employ GPT-4o, GPT-3.5-turbo, and Gemini 2.5, which are state-of-the-art models in reasoning tasks. Each model evaluates the responses independently in two separate runs, resulting in a total of six scores per response. We report the average of all scores as the final result. This evaluation strategy enables large-scale, consistent assessment of qualitative response quality, bypassing the need for labor-intensive human annotation.


\vspace{1ex}
\noindent \textbf{System performance metrics.} 
To evaluate the efficiency of \system{}, we measure the end-to-end latency, defined as the time elapsed from receiving the user's input to the generation of the final response token. Additionally, we report the total token length, calculated by summing the output token length of the front-end model (i.e., the expert LLM in \system{}) and both the input and output token lengths of the back-end model (i.e., the response-generation LLM). This total includes the system prompt but excludes any padding or internal tokens used purely for generation. Token length is a critical efficiency metric, as it directly affects inference latency and computational cost, with longer input/output sequences increasing generation time and system resource usage.

\subsection{End-to-end Response Quality Performance}
\label{subsec:quality}

As the first part of our evaluation, we assess the overall performance of \system{} based on three response quality metrics. We begin by evaluating \system{} as a complete end-to-end framework to establish a baseline for its effectiveness in generating high-quality responses.


\begin{figure}[t]
    \centering
    \includegraphics[width=.8\linewidth]{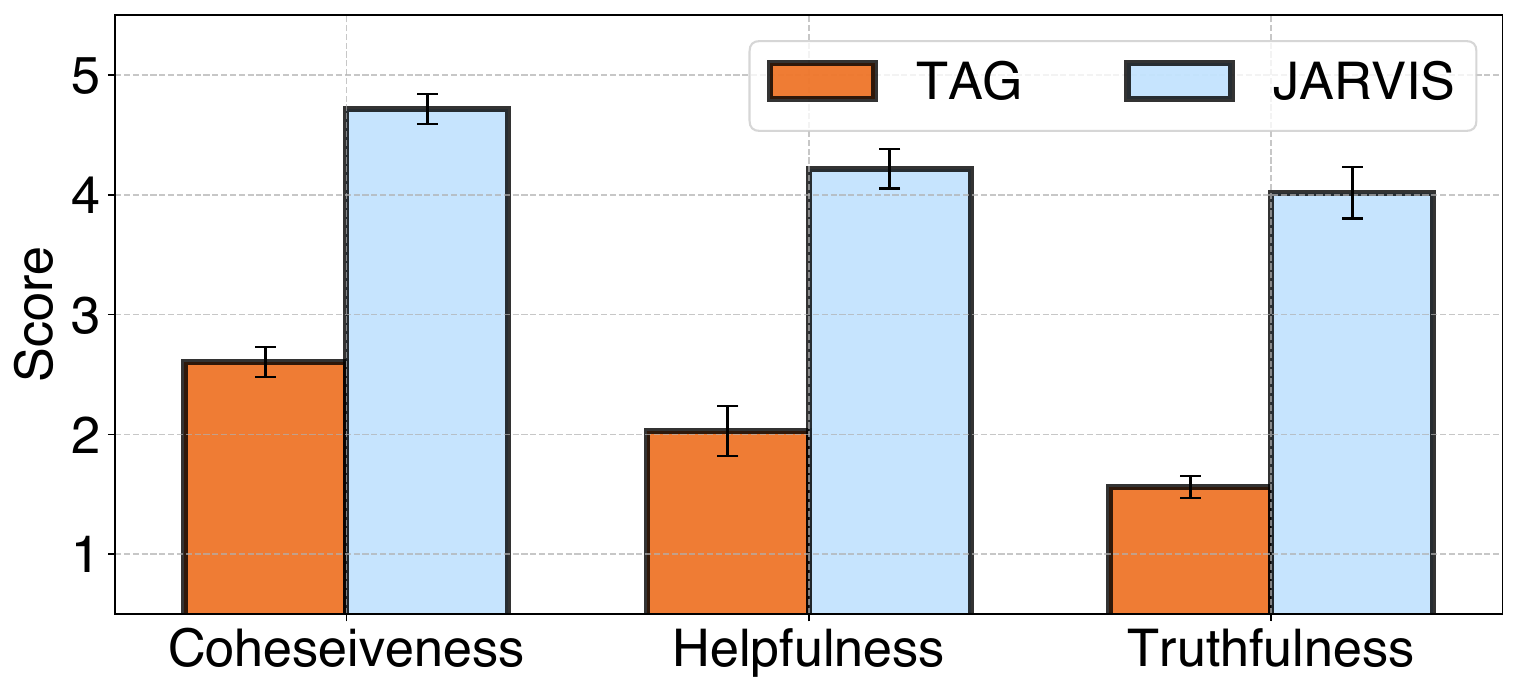}
    \vspace{-2ex}
    \caption{Response quality comparison vs Baseline TAG~\cite{biswal2024text2sql}.}
    \vspace{-2ex}
    \label{fig:baseline_comparison}
\end{figure}

Figure~\ref{fig:baseline_comparison} presents the overall performance of \system{} across the three response quality metrics evaluated in this study. We compare \system{} against the RAG-based framework proposed by Biswal et al., known as TAG~\cite{biswal2024text2sql}, which serves as our baseline. To the best of our knowledge, no prior work has explored LLM-based RAG for sensor-data-driven HVAC QA. Thus, TAG, which designed as a general-purpose framework for LLM-database interaction, offers the most suitable point of comparison. All scores reported in Figure~\ref{fig:baseline_comparison} are obtained using the LLM-as-a-Judge evaluation methodology.

As the results show, \system{} outperforms TAG across all three response quality metrics. This performance gap is primarily stems from TAG’s limited ability to incorporate HVAC-specific context and its susceptibility to erroneous SQL generation, which occurred in 4 out of 27 cases. In fact, these findings highlight the importance of employing an expert LLM at the front end of the system, as it enables \system{} to more effectively interpret and calibrate user inputs within the HVAC domain. Additionally, in three instances, TAG successfully retrieved relevant data but failed to produce a response because the data exceeded the input context limit of the response-generation LLM.

In contrast, \system{} achieved average scores above 4.0 in both \textit{cohesiveness} and \textit{truthfulness}, and 3.7 in \textit{helpfulness}. Based on the LLM judges’ reasoning and a deeper analysis, we observe that \system{} demonstrates strong capabilities in interpreting user intent through \textit{semantic context injection} at the Expert-LLM, resulting in coherent and well-aligned answers. Additionally, the integration of structured data processing, performed at the Data processing module, enhances factual accuracy by enabling advanced statistical computations, which contributes to higher truthfulness scores.

However, we also identified limitations in cases where the query results deviate from typical expectations. For instance, when asked to compare two temperature values that happen to be identical, a user would expect the system to respond that the values are ``the same.'' Instead, \system{} returned a rigid response stating that one value was ``larger by 0.'' In such cases, the system fails to adapt its output flexibly, reducing the perceived \textit{helpfulness} of the response. This highlights a limitation where \system{} prioritizes factual correctness but lacks contextual sensitivity and the ability to deliver responses that are more natural, concise, and user-aligned.

\subsection{Component Level Response Quality}
\label{subsec:eval_component}
\begin{figure}[t]
    \centering
    \includegraphics[width=.8\linewidth]{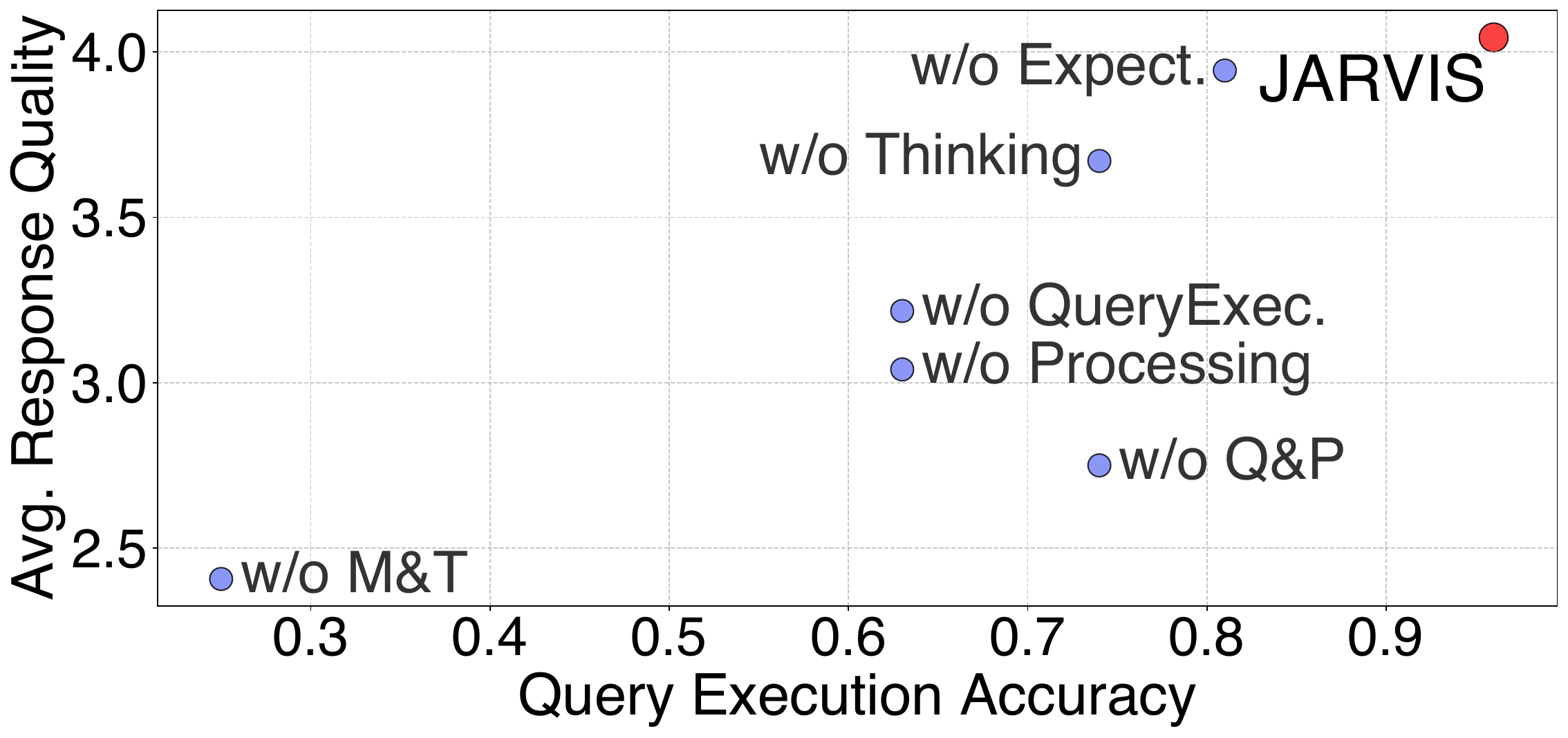}
    \vspace{-2ex}
    \caption{Avg. response quality vs. Query execution accuracy.}
    \vspace{-2ex}
    \label{fig:queryem-tradeoff}
\end{figure}

\begin{figure}[t]
    \centering
    \includegraphics[width=.8\linewidth]{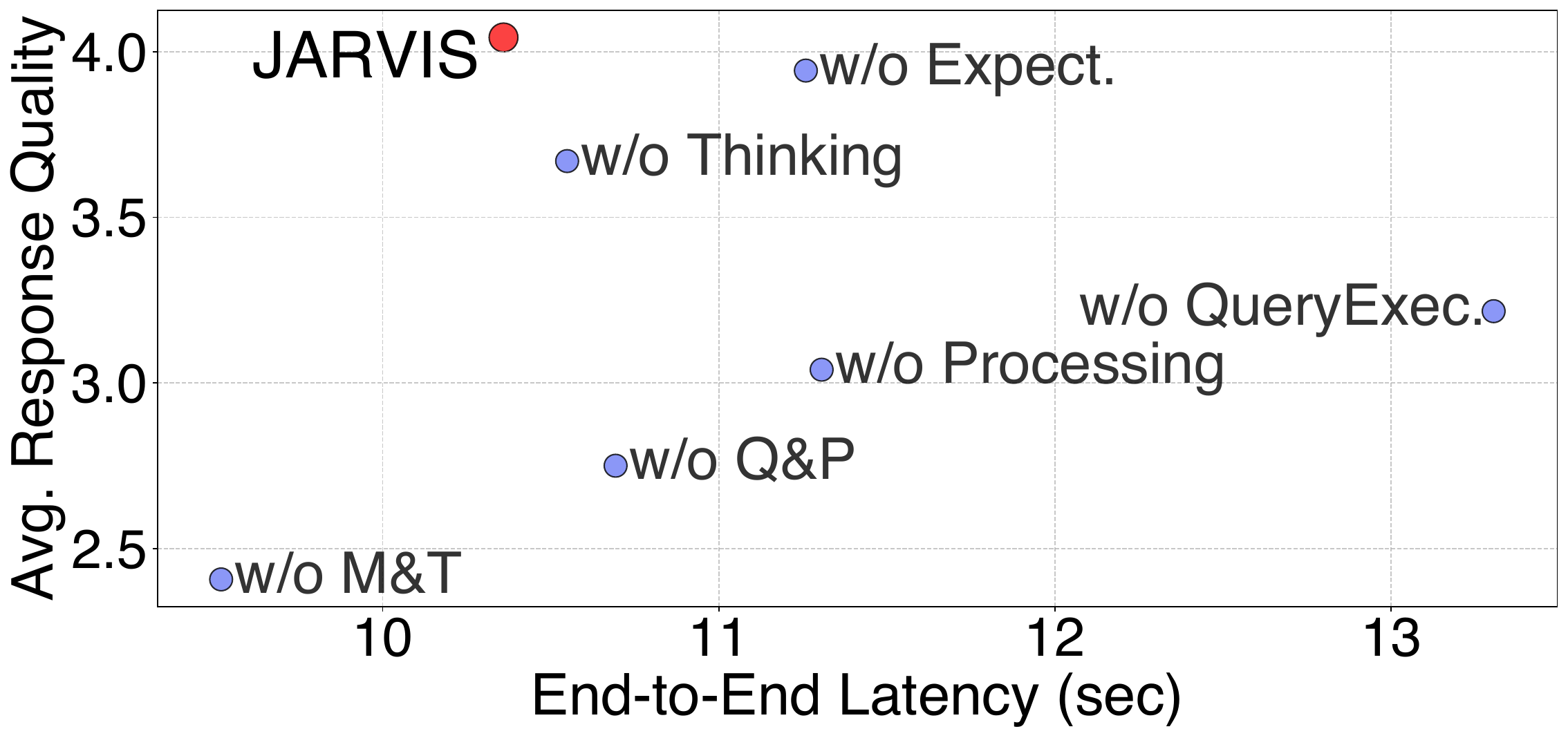}
    \vspace{-2ex}
    \caption{Avg. response quality vs. End-to-end latency.}
    \vspace{-2ex}
    \label{fig:overall-tradeoff}
\end{figure}

Given the end-to-end performance of \system{}, we now shift focus to a more fine-grained, component-level analysis of response quality. The goal of this analysis is to assess the effectiveness of each core design choice in \system{} and understand how individual components contribute to overall system performance.

Figure~\ref{fig:queryem-tradeoff} presents the results of this analysis by comparing \system{} against several ablation variants, where one or more key modules are removed. This comparison highlights the benefits of \system{}'s architecture and quantifies the contribution of each component to average response quality.

We evaluate seven configurations in total. First, we remove both Metadata injection and the Thinking module from the Expert-LLM ({\tt w/o~M\&T}), simulating a generic model with no domain adaptation or tailored reasoning capabilities. Second, we reintroduce Metadata while omitting the Thinking module ({\tt w/o~Thinking}), reflecting a configuration with contextual data but no specialized reasoning. Third, we test a setup with Metadata and Thinking enabled, but with the Expectation module removed ({\tt w/o~Expect.}), which prevents the system from aligning instructions with inferred user goals.

Turning to the Agent component, we remove the Parameterized Query Execution module ({\tt w/o~QueryExec.}), responsible for translating instructions into robust SQL queries. We then remove the Data Processing module ({\tt w/o~Processing}), assessing performance when statistical operations rely solely on native SQL. We also examine a configuration with both of these modules removed ({\tt w/o~Q\&P}), resulting in a minimal backend. All variants are compared against the full implementation of \system{}.



\begin{figure*}[t]
    \centering
    \includegraphics[width=.80\linewidth]{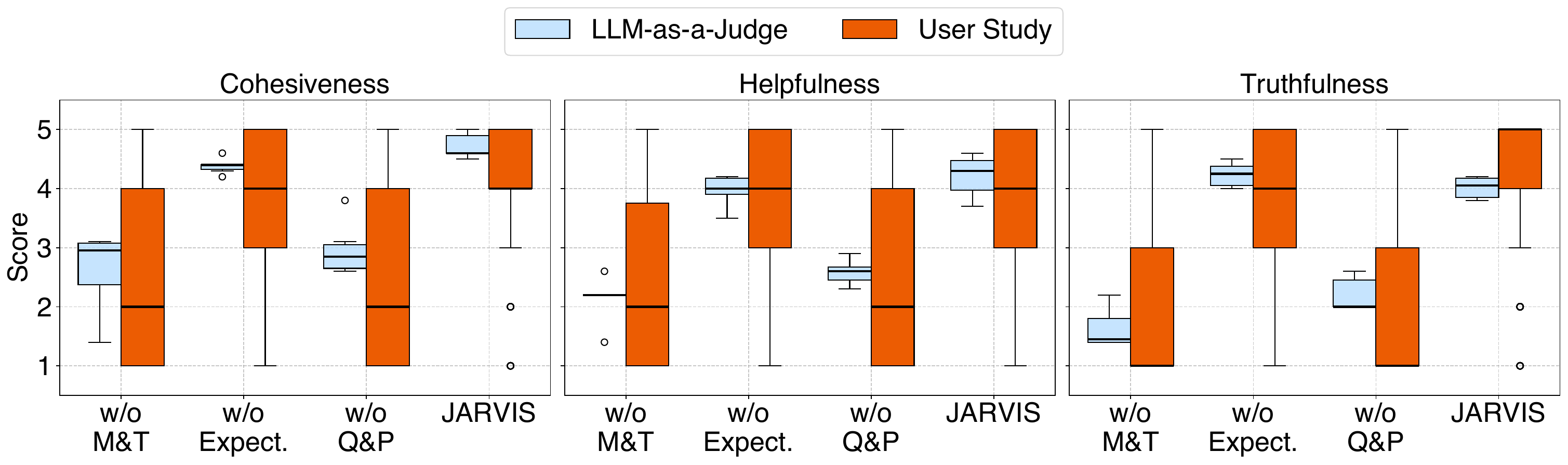}
    \vspace{-3ex}
    \caption{Response quality measure comparison between the LLM-as-a-Judge approach and user study. }
    \vspace{-2ex}
    \label{fig:laj-userstudy-alignment}
\end{figure*}

As Figure~\ref{fig:queryem-tradeoff} shows, \system{}, equipped with all components, consistently outperforms its ablation variants in response quality. Notice that while query accuracy and response quality are generally well-aligned, we see some notable exceptions. For instance, the {\tt w/o~Q\&P} configuration yields comparable (and occasionally higher) query match accuracy compared to the overall linear trend, but suffers from markedly lower response quality. Similarly, while {\tt w/o~QueryExec} and {\tt w/o~Processing} achieve similar query accuracy, {\tt w/o~QueryExec} consistently outperforms {\tt w/o~Processing} in response quality. These findings reinforce our claim that accurate data retrieval alone is insufficient and LLMs require external statistical processing to reason effectively over structured data. \system{} addresses this through its integrated data processing module.

We further observe a super-additive synergy when both the Query Execution and Processing modules are used together. Ablating either component leads to noticeable performance drop, but combining both (c.f., full \system{}) achieves notable improvement. Further analysis reveals that the Query Execution module produces deterministic SQL queries with predictable result schemas, while the Processing module generates Python logic tailored to well-consume these schemas. This alignment reduces execution errors and boosts end-to-end robustness.


A similar pattern holds for {\tt w/o~M\&T}, {\tt w/o~Thinking}, and the full \system{} configuration. Metadata injection improves both query accuracy and response quality by enriching user intent with contextual information. Adding the Thinking component further enhances performance by enabling structured reasoning over ambiguous or underspecified queries. These results confirm the architectural value of semantic grounding and intermediate reasoning in \system{}.

In addition, Figure~\ref{fig:overall-tradeoff} presents the trade-off between response quality and end-to-end inference time. Interestingly, \system{}, which achieves the highest response quality, also ranks as the second-fastest configuration among all variants. While one might expect that omitting the Thinking component (i.e., {\tt w/o~M\&T} and {\tt w/o~Thinking}) would reduce inference time due to fewer tokens being generated, we find a counter-intuitive result with the full \system{} that includes the Thinking component being slightly faster than {\tt w/o~Thinking}. This is because the Thinking component enables more optimized execution paths by planning efficient queries and processing steps, which compensates for the additional token cost during generation. A similar benefit is observed with the Expectation module, which helps guide the system toward efficient and focused responses.

Overall, these results demonstrate that \system{} is a performance-efficient framework that achieves both high response quality and low inference latency through principled modular design.

\vspace{-2ex}
\subsection{Response Quality User Study}

While the LLM-as-a-Judge framework provides a scalable and cost-effective evaluation for LLM outputs, it remains an automated proxy. To complement this, we conducted a user study to (i) validate the reliability of automated scores and (ii) capture qualitative aspects of user experience that may be missed by automatic evaluation.

We carried out an IRB-approved user study on a carefully selected subset of the test set. The study comprised 10 `user query'-`system response' pairs generated from different \system{} configurations, with participants rating individual responses on a 1-5 Likert scale across three response quality metrics: \textit{cohesiveness}, \textit{helpfulness}, and \textit{truthfulness}. Each participant was provided with a short scenario description and a sample evaluation to calibrate their understanding of the persona and rubric. The 10 questions span the full spectrum of query categories in our dataset, as illustrated in Figure~\ref{fig:data-distribution}. The study involved 22 participants (avg age: 35.9, stdev: 11.0). Among them, 45.7\% reported daily use of HVAC systems with a good understanding of their operation, and 8.6\% were professionals in HVAC-related industries. In terms of LLM familiarity, 45.5\% indicated moderate experience, while 24.2\% reported using LLMs primarily for question answering and information search.


\begin{figure*}[t]
    \centering
    \includegraphics[width=.80\linewidth]{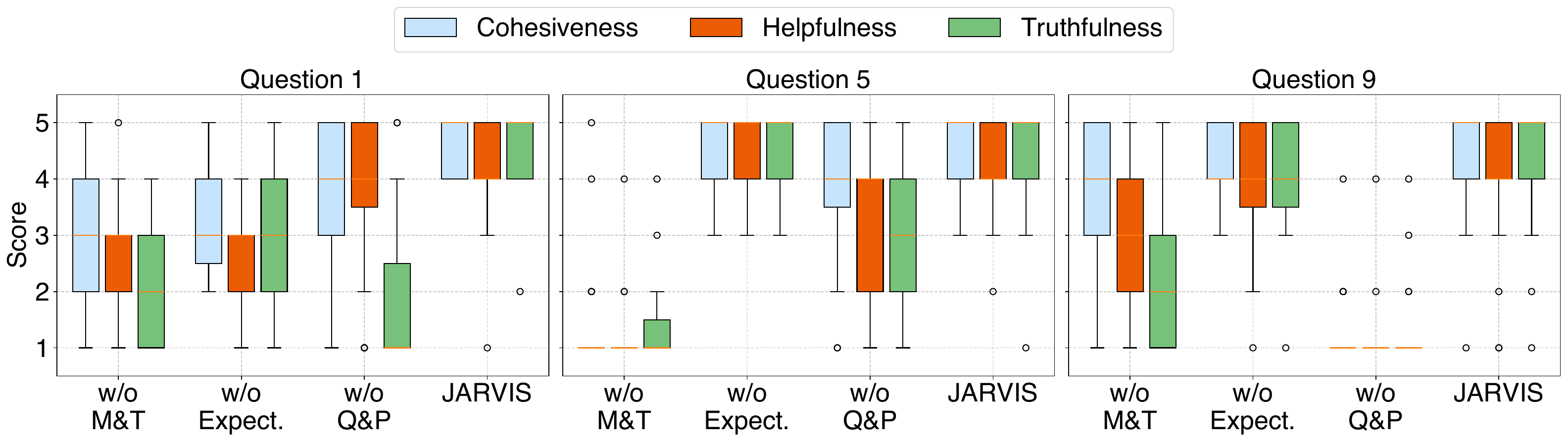}
    \vspace{-2ex}
    \caption{User study results for Questions 1, 5 and 9. Error bars indicate 25th–75th percentile range.}
    \vspace{-2ex}
    \label{fig:userstudy-perquestion}
\end{figure*}


Figure~\ref{fig:laj-userstudy-alignment} compares the average response quality scores produced by the LLM-as-a-Judge with those obtained from the user study on the same 10 data points. While user ratings exhibit higher variance, the LLM-based scores closely track the median of the human evaluations. Moreover, the relative ranking of the ablation variants remains highly consistent across both evaluation methods, suggesting that the automated LLM-as-a-Judge framework serves as a reliable proxy for human assessment within our dataset and application scenario. Consistent with earlier findings, the full implementation of \system{} achieves the highest ratings across all three response quality metrics, indicating that its responses are well-received in all dimensions. This highlights the robustness and overall reliability of \system{} across diverse query scenarios.


We further analyze the user study on a per-question basis to gain deeper insight into individual queries and corresponding user responses. The query presented in Question 1 of the study was: \textit{``What is this week's average temperature in my room and the room next door?''}, a query posed by a building manager's persona. Study participants were asked to rate the responses generated by different system configurations using the same response quality metrics. As the left plot in Figure~\ref{fig:userstudy-perquestion} shows, \system{} is the only configuration that consistently receives high scores across all three metrics. Even the {\tt w/o~Expect} variant, which had shown competitive performance in earlier evaluations, underperforms on this question.

A notable observation is the low performance of the {\tt w/o~M\&T} configuration, primarily due to its failure to generate valid SQL queries when faced with ambiguous references like ``my room'' and ``room next door.'' In contrast, \system{} leverages metadata injection and the semantic reasoning of the Thinking module to resolve such ambiguities, allowing the Expert-LLM to produce contextually grounded and deployment-specific queries.

In the case of the {\tt w/o~Q\&P} variant, we observe that while the system correctly interprets the user's intent and generates a plausible answering plan, it ultimately fails to compute an accurate result due to the absence of statistical processing. This leads to a significant drop in factual accuracy, implying the importance of the Agent’s Query Execution and Processing modules in supporting mathematically sound and reliable answers.

Surprisingly, in the {\tt w/o~Expect} configuration, the Expert-LLM correctly interpreted the user's intent and outlined a reasonable answering pipeline. However, it produced execution instructions that misaligned with the high-level plan generated by the Thinking component. Specifically, instead of computing separate weekly averages for ``my room'' and ``the room next door,'' the system averaged across both rooms in a single computation, conflating the intended outputs. This misalignment led to lower cohesiveness and helpfulness scores, as the response failed to meet the user's implied expectations. In contrast, the Expectation module mitigated this issue by explicitly defining the response goal via a reference answer format, which specified two distinct values. This goal-setting step guides subsequent instruction generation toward the intended structure, enhancing both clarity and user alignment. We refer to this design, where the response goal is first set and then used to inform downstream planning, as an instance of \textit{bottom-up planning}.

In Question 5, a resident persona asks: \textit{``When was the hottest day two weeks ago?''} We classify this query as one with implicit and ambiguous conditions. The Expert-LLM is expected to infer that ``hottest'' refers to the highest indoor temperature, and that the absence of spatial information implies the user is referring to their own room. We consider these inferences part of HVAC-common knowledge. As expected, the {\tt w/o~M\&T} case exhibits low response quality, as it lacks contextual reasoning and the domain-specific adaptation required for interpreting HVAC terminology and intent.

In addition, when data processing support is removed (i.e., {\tt w/o~Q\&P}), the system fails to correctly compute the \texttt{argmax} operation needed to determine the hottest day. This is because \texttt{argmax} is not natively supported by SQL and requires additional computation through the processing module. Without it, the system's truthfulness score drops significantly due to factual inaccuracies in the response.

In Question 9, a resident asks: \textit{``What was the highest and lowest room temperature in the last summer?''} This question evaluates the system’s ability to handle ambiguous temporal expressions; specifically, interpreting the term ``summer'' as a concrete date range. Baseline general-purpose LLMs often struggle with this interpretation, and we designed this question to test whether the Expert-LLM can generate meaningful temporal constraints. 

With the Thinking component missing (i.e., {\tt w/o~M\&T}), the system captures the user’s overall intent but fails to resolve “summer” into specific dates.  
Since the Thinking module is trained to support reasoning by mapping generic user expressions to HVAC-relevant contexts, its absence leads to ambiguity in interpreting temporal references. As a result, configurations equipped with the Thinking module are able to estimate a reasonable summer timeframe and successfully incorporate it into the answering pipeline. Nevertheless, in the {\tt w/o~Q\&P} configuration, the system fails to translate the temporal condition into raw SQL due to the absence of a specialized query generation module, again highlighting the importance of the Agent’s back-end components in supporting HVAC-specific logic.






\subsection{System-level Query Accuracy Analysis}


\begin{figure}[t]
    \centering
    \includegraphics[width=0.95\linewidth]{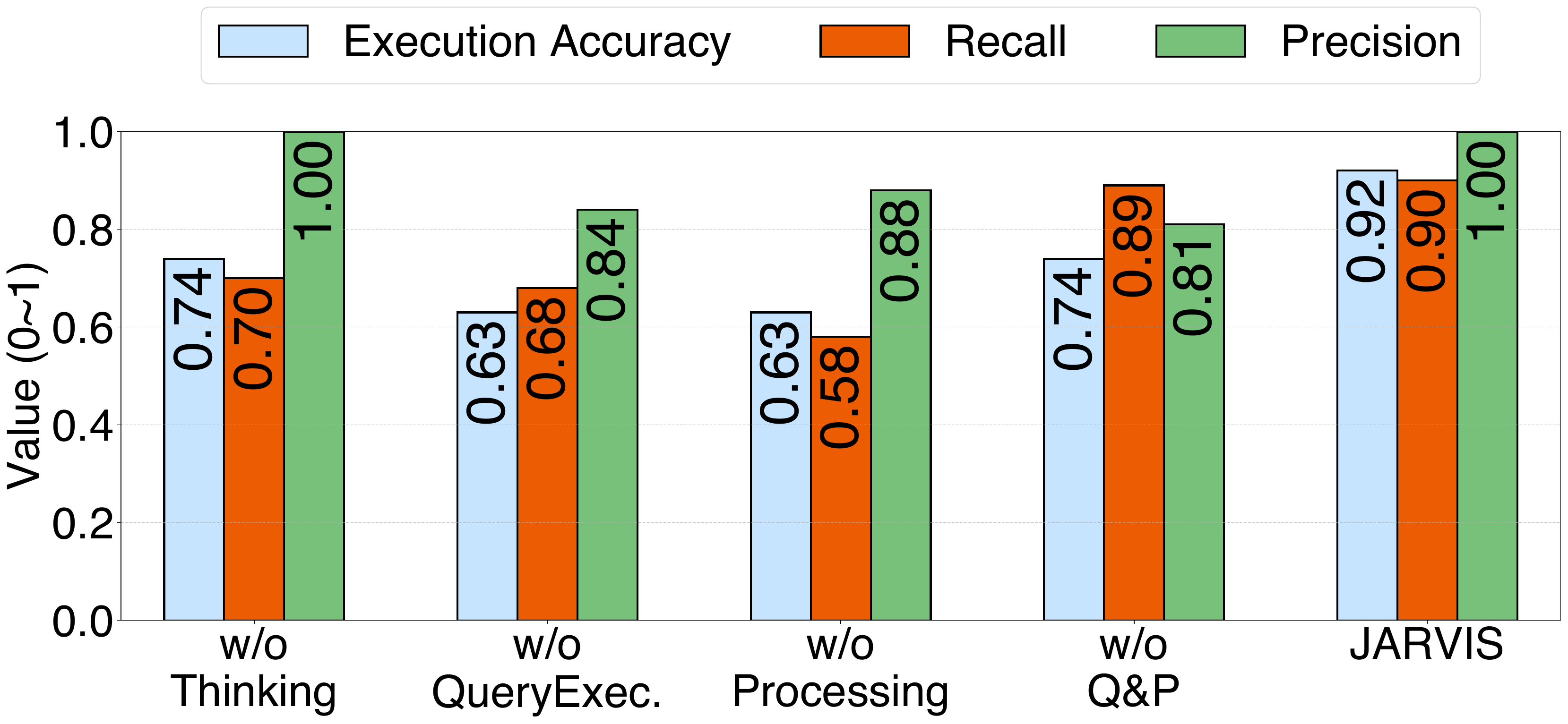}
    \vspace{-3ex}
    \caption{Query execution accuracy, recall and precision for different \system{} variants.}
    \vspace{-3ex}
    \label{fig:db_performance}
\end{figure}

As briefly discussed in Section~\ref{subsec:eval_component}, query accuracy tends to show a positive correlation with the overall response quality. In this section, we conduct a deeper analysis of how each component in \system{} affects data querying performance by decomposing the query accuracy metric into precision and recall.





With the full \system{} case showing the best overall performance, Figure~\ref{fig:db_performance} provides a deeper view of the synergy between the Query Execution and Processing modules, as previously observed when analyzing Figure~\ref{fig:queryem-tradeoff}. Compared to {\tt w/o~Q\&P}, enabling both modules (full \system{}) significantly improves precision, recall, and accuracy. When only one is present ({\tt w/o~QueryExec} or {\tt w/o~Processing} compared to {\tt w/o~Q\&P}), the recall shows a sharp drop while precision remains stable, suggesting that each module acts as an \textit{aggressive filter} with the Query Executor achieving this through stricter SQL constraints, and the Processing module via post-query logic. These filters operate from distinct angles, and when combined, their complementary behaviors yield a \textit{paradoxical synergy} that enhances both selectivity and coverage, thereby improving response quality.


Furthermore, we argue that the aggressiveness of each module stems from their distinct operational characteristics. Comparing the {\tt w/o~QueryExec} and {\tt w/o~Processing} cases reveals a trade-off between recall and precision under similar accuracy. This suggests that the Query Execution module functions as a more aggressive filter. We attribute this behavior to \system{}'s design philosophy of decomposing large, monolithic SQL queries into smaller, modular sub-queries. This modularization enables more precise and repeated querying, inherently favoring high-precision execution at the cost of recall; thus, reinforcing its role as an aggressive filter.

Another interesting observation from Figure~\ref{fig:db_performance} regards the Thinking component. Even without this module ({\tt w/o~Thinking}), the system achieves precision comparable to the full \system{} configuration. This indicates that the Thinking component primarily contributes to improvements in accuracy and recall by enforcing semantically grounded reasoning and enhancing the system’s ability to retrieve relevant but potentially under-specified data.

\subsection{Processing Latency Breakdown}

\begin{table}[t]
\centering
\begin{adjustbox}{width=\linewidth} 
\begin{tabular}{l|c|c|c|c}
\hline
Latency (sec) & \textbf{Expert LLM}   & \textbf{Querying}& \textbf{Processing}& \textbf{Response} \\ \hline\hline
Proc. X& 5.47 ± 0.69& 0.00 ± 0.00      & 0.24 ± 0.01 & 0.39 ± 0.08\\ \hline
Proc. O& 8.86 ± 4.74& 0.00 ± 0.01 & 1.73 ± 2.51 & 0.57 ± 0.39\\\hline \hline
Ambig. X& 6.17 ± 2.53& 0.01 ± 0.02 & 0.59 ± 0.78 & 0.55 ± 0.49\\\hline
Ambig. O& 9.16 ± 4.88& 0.00 ± 0.00 & 1.85 ± 2.67 & 0.53 ± 0.31\\\hline
\end{tabular}
\end{adjustbox}
\caption{Latency observed at each processing module within \system{} for different types of input query requests.}
\vspace{-4ex}
\label{tab:execution_time}
\end{table}

\begin{figure}[t]
    \centering
    \includegraphics[width=0.95\linewidth]{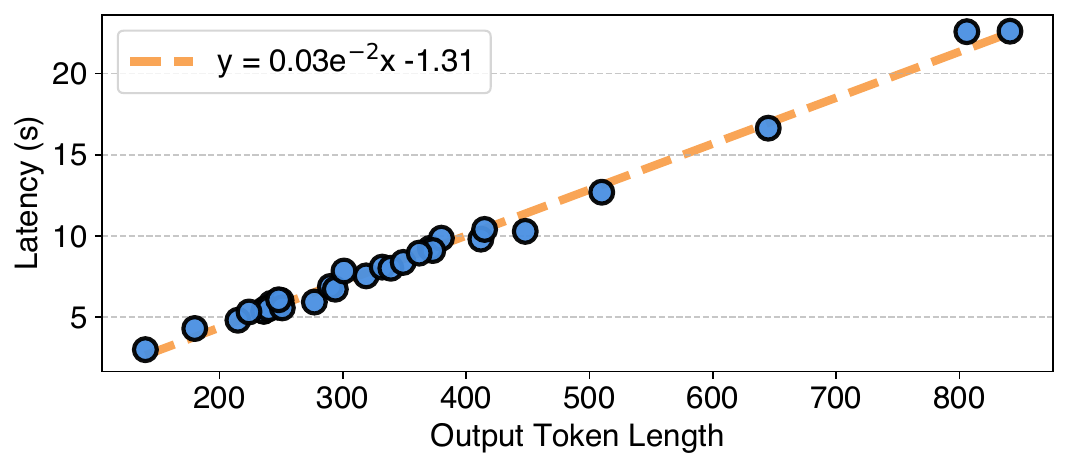}
    \vspace{-4ex}
    \caption{Expert-LLM latency for varying token lengths.}
    \vspace{-2ex}
    \label{fig:latency}
\end{figure}

\system{} comprises several software components that together form an end-to-end pipeline from user input to final response. While there is no strict latency requirement, the system must respond within a timeframe deemed ``acceptable'' for practical deployment.

In Table~\ref{tab:execution_time}, we present the latency incurred by each component of \system{} for various queries involving different operations. The most notable observation is that the Expert-LLM accounts for the highest latency among all components, significantly more than the response-generating LLM. This is primarily because the Expert-LLM serves as the entry point to \system{}, processing the full prompt and user input, which typically involves longer token sequences. In contrast, the response-generation LLM operates on more structured and concise inputs passed from other components within the Agent.

Queries that require some level of processing ({\tt Proc. O} in the table) exhibit higher latency at both the Expert-LLM and the data processing stage compared to those that do not ({\tt Proc. X}). While longer data processing time is expected, the increased latency at the Expert-LLM stems from the need to generate more complex internal execution instructions to support downstream Agent operations. A similar pattern is observed for queries with and without ambiguous terms (i.e., {\tt Ambig. O} and {\tt Ambig. X}), as input ambiguity typically necessitates more elaborate reasoning and planning by the Expert-LLM. Additionally, the processing latency is higher for {\tt Ambig. O}, which we attribute to our QA dataset containing more complex processing tasks for ambiguous queries.

We further analyze the Expert-LLM’s latency with respect to its output token length in Figure~\ref{fig:latency}. As the results show, there is a positive correlation between latency and token length, indicating that complex queries requiring longer and detailed execution instruction sets can lead to increased latency at the Expert-LLM.

While our LLMs offer a favorable balance between quality and latency suitable for our use case, we acknowledge the opportunity for future improvements. Exploring smaller, optimized model architectures (via quantization, distillation, or Mixture-of-Experts designs) may further reduce inference time without compromising performance. We leave this exploration to future work.

\section{Discussions}

Based on our experience designing and evaluating \system{}, we outline several discussion points that highlight limitations and open opportunities for future research.

\vspace{1ex}\noindent\textbf{Failure cases and schema conformance.}
A recurring challenge we observed involves the Expert-LLM's generation of malformed JSON outputs, despite targeted prompt tuning and fine-tuning efforts. These errors, such as missing or extraneous brackets, result in schema violations that cause execution failures. While relatively infrequent (1 out of 27 test cases in our evaluation), they highlight a broader issue when relying on free-form text generation for structured outputs. In this work, we address the problem using a regex-based post-processing patch; however, this approach remains brittle and reactive. An interesting direction for future work is to incorporate structured decoding mechanisms (e.g., constrained or programmatic decoding) to ensure that outputs conform to expected schemas by construction. This could improve system robustness and reduce downstream error handling complexity.

\vspace{1ex}
\noindent \textbf{Expert-LLM as a semantic context manager.}
The Expert-LLM in \system{} serves as a centralized module for acquiring and propagating domain-specific semantic context to downstream components. This centralized architecture offers broader applicability beyond HVAC, especially in modular frameworks that combine expert LLMs with other specialized components (including additional LLMs). By offloading semantic interpretation to a single module, the system reduces the need to re-engineer context representations across modules, lowering development overhead, improving scalability, and simplifying maintenance.

\vspace{1ex}\noindent\textbf{LLM architecture flexibility.} While we do not explore alternative LLM architectures in this work, the core design of \system{} is modular and can accommodate different models for both the Expert-LLM and the response generation LLM. For the scope of this work, we selected {\tt LLaMA3.1-8B-Instruct} and {\tt LG EXAONE-3.5-7.8B} to ensure that \system{} can operate as a stand-alone, on-site system with strong support for the native language used in our deployment. Exploring other LLM architectures, tailored for different deployment constraints, languages, or HVAC use cases, remains a promising direction for future research and wide-spread deployment.


\vspace{1ex}
\noindent \textbf{Trade-off between response richness and stability.} As briefly discussed in Section~\ref{subsec:quality}, \system shows a key limitation in response richness given that the response generation LLM is inclined to focus on the expectation template. While this bias reliably yields context-rich answers even from minimal semantic contexts, it can lessen the naturalness of the responses by curbing the model's expressive freedom. In essence, this is a trade-off between producing flexible, conversational responses and delivering information-dense answers with lower risk of omission or hallucination, which is an intentional design choice we made to favor the latter. Balancing these competing objectives, perhaps by fine-tuning the response generator, is left for future work.

\section{Conclusion}

This work presented \system{}, a modular LLM-based framework designed to enable accurate and context-aware question answering in HVAC environments. By integrating structured sensor data with semantic reasoning through components such as metadata injection, intermediate thought generation, and agent-side processing, \system{} overcomes key limitations of traditional LLMs in dynamic, sensor-rich domains. Our evaluations, which included automated scoring via LLM-as-a-Judge, a real-world user study, and system performance analysis, demonstrates that \system{} significantly outperforms baseline and ablation variants across all key response quality metrics. Notably, the results show that each system component contributes meaningfully to both response quality and execution efficiency. Furthermore, our analysis reveals that \system{} is capable of providing accurate and helpful responses even under underspecified or ambiguous conditions, showcasing its practical value in real-world deployments.



\balance
\bibliographystyle{ACM-Reference-Format}
\bibliography{references}

\clearpage

\end{document}